\documentclass{article}

\usepackage{microtype}
\usepackage{graphicx}
\usepackage{subfigure}
\usepackage{listings}
\usepackage{booktabs} 
\usepackage{comment}
\usepackage{algorithm}
\usepackage{algorithmic}
\usepackage{url}

\usepackage{breakurl}
\usepackage[breaklinks]{hyperref}

\usepackage{xcolor} 
\usepackage{tikz}
\usetikzlibrary{fit, calc}
\newcommand*{\tikzmk}[1]{\tikz[remember picture, overlay] \node (#1) {};\ignorespaces}
\newcommand{\boxit}[1]{\tikz[remember picture, overlay]{\node [yshift=-9pt, xshift=-17pt, fill=#1, opacity=.25, fit={(A)($(B) + (1.07\linewidth, -0.2\baselineskip)$)}] {};}\ignorespaces}
\colorlet{pink}{red!40}

\usepackage{amsmath}

\DeclareMathOperator*{\argmax}{arg\,max}

\usepackage{comment}

\usepackage[accepted]{icml2020}

\icmltitlerunning{Discriminative Adversarial Search for Abstractive Summarization}

\begin{document}

\twocolumn[
\icmltitle{Discriminative Adversarial Search for Abstractive Summarization}



\icmlsetsymbol{equal}{*}

\begin{icmlauthorlist}
\icmlauthor{Thomas Scialom}{recital,lip6}
\icmlauthor{Paul-Alexis Dray}{recital}
\icmlauthor{Sylvain Lamprier}{lip6}
\icmlauthor{Benjamin Piwowarski}{lip6,cnrs}
\icmlauthor{Jacopo Staiano}{recital}
\end{icmlauthorlist}

\icmlaffiliation{recital}{reciTAL, Paris, France}
\icmlaffiliation{cnrs}{CNRS, France}
\icmlaffiliation{lip6}{Sorbonne Universit\'e, CNRS, LIP6, F-75005 Paris, France}

\icmlcorrespondingauthor{Thomas Scialom}{thomas@recital.ai}

\icmlkeywords{Machine Learning, Natural Language Generation, Summarization, ICML}

\vskip 0.3in
]



\printAffiliationsAndNotice{}  
\begin{abstract} 
We introduce a novel approach for sequence decoding, \emph{Discriminative Adversarial Search} (DAS), which has the desirable properties of alleviating the effects of exposure bias without requiring external metrics. Inspired by Generative Adversarial Networks (GANs), wherein a discriminator is used to improve the generator, our method differs from GANs in that the generator parameters are not updated at training time and the discriminator is only used to drive sequence generation at inference time. 
We investigate the effectiveness of the proposed approach on the task of Abstractive Summarization: the results obtained show that  DAS improves over the state-of-the-art methods, with further gains obtained via discriminator retraining. Moreover, we show how DAS can be effective for cross-domain adaptation. Finally, all results reported are obtained without additional rule-based filtering strategies, commonly used by the best performing systems available: this indicates that DAS can effectively be deployed without relying on post-hoc modifications of the generated outputs.
\end{abstract}

\section{Introduction}
\label{section:introduction}
In the context of Natural Language Generation (NLG), a majority of approaches propose sequence to sequence models trained via maximum likelihood estimation; a Teacher Forcing \cite{williams1989learning} strategy is applied during training: ground-truth tokens are sequentially fed into the model to predict the next token.
Conversely, at inference time, ground-truth tokens are not available: the model can only have access to its previous outputs. In the literature \cite{bengio2015scheduled, ranzatosequence}, such mismatch is referenced to as \emph{exposure bias}: as mistakes accumulate, this can lead to a divergence from the distribution seen at training time, resulting in poor generation outputs.

Several works have focused on alleviating this issue, proposing to optimize a \emph{sequence level metric} such as BLEU or ROUGE: \citet{wiseman2016sequence} used beam search optimisation while \citet{ranzatosequence} framed text generation as a reinforcement learning problem, using the chosen metric as reward. Still, these automated metrics suffer from known limitations: \citet{sulem-etal-2018-bleu} showed how BLEU metrics do not reflect meaning preservation, while \citet{novikova-etal-2017-need} pointed out that, for NLG tasks, they do not map well to human judgements. 

Similar findings have been reported for ROUGE, in the context of abstractive summarization \cite{paulus2017deep}: for the same input, several correct outputs are possible; nonetheless, the generated output is often compared to a single human reference, given the lack of annotated data. Complementary metrics have been proposed to evaluate NLG tasks, based on Question Answering \cite{scialom2019answers} or learned from human evaluation data \cite{bohm2019better}. Arguably, though, the correlation of such metrics to human judgments is still unsatisfactory.

To tackle exposure bias, Generative Adversarial Networks (GANs) \cite{goodfellow2014generative} represent a natural alternative to the proposed approaches: rather than learning from a specific metric, the model learns to generate text that a discriminator cannot differentiate from human-produced content. 
However, the discrete nature of text makes the classifier signal non-differentiable. A solution would be to use reinforcement learning with the classifier prediction as a reward signal. However, due to reward sparsity and mode collapse \cite{Zhou2020Self-Adversarial}, text GANs failed so far to be competitive with state-of-the-art models trained with teacher forcing on NLG tasks \cite{caccia2018language, clark2019electra}, and are mostly evaluated on synthetic datasets.

Inspired by Generative Adversarial Networks, we propose an alternative approach for sequence decoding: first, a discriminator is trained to distinguish human-produced texts from machine-generated ones. Then, this discriminator is integrated into a beam search: at each decoding step, the generator output probabilities are refined according to the likelihood that the candidate sequence is human-produced. This is equivalent to optimize the search for a custom and dynamic metric, learnt to fit the human examples. 

Under the proposed paradigm, the discriminator causes the output sequences to diverge from those originally produced by the generator. These sequences, adversarial to the discriminator, can be used to further fine-tune the discriminator:
following the procedure used for GANs, the discriminator can be iteratively trained on the new predictions it has contributed to improve. 
This effectively creates a positive feedback loop for training the discriminator: until convergence, the generated sequences improve and become harder to distinguish from human-produced text.
Additionally, the proposed approach allows to dispense of custom rule-based strategies commonly used at decoding time such as length penalty and n-gram repetition avoidance.

In GANs, the discriminator is used to improve the generator and is dropped at inference time. Our proposed approach differs in that, instead, we do not modify the generator parameters at training time, and use the discriminator at inference time to drive the generation towards human-like textual content.

The main contributions of this work can be summarized as:
\begin{enumerate}
    \item we propose \emph{Discriminative Adversarial Search} (DAS), a novel sequence decoding approach that allows to alleviate the effects of exposure bias and to optimize on the data distribution itself rather than for external metrics;
    \item we apply DAS to the abstractive summarization task, showing that even without the self-retraining procedure, our 
    discriminated beam search procedure  improves over the state-of-the-art for various metrics;
    \item we report further significant improvements when applying discriminator retraining;
    \item finally, we show how the proposed approach can be effectively used for domain adaptation. 
\end{enumerate}

\section{Related Work}
\subsection{Exposure Bias}
Several research efforts have tackled the issue of exposure bias resulting from Teacher Forcing.
Inspired by \citet{venkatraman2015improving}, \citet{bengio2015scheduled} proposed a variation of Teacher Forcing wherein the ground truth tokens are incrementally replaced by the predicted words. Further, Professor Forcing \cite{lamb2016professor} was devised as an adversarial approach in which the model learns to generate without distinction between training and inference time, when it has no more access to the ground truth tokens. Using automated metrics at coarser (sequence) rather than finer (token) granularity to optimize the model, \citet{wiseman2016sequence} proposed a beam search variant to optimise the BLEU score in neural translation. Framing NLG as a Reinforcement Learning problem, \citet{ranzatosequence} used the reward as the metric to optimise. \citet{paulus2017deep} applied a similar approach in abstractive summarization tasks, using the ROUGE metric as a reward; the authors observed that, despite the successful application of reinforcement, higher ROUGE does not yield better models: other metrics for NLG are needed. Finally, \citet{zhang2019bridging} proposed to select, among the different beams decoded, the one obtaining the highest BLEU score and then to fine-tune the model on that sequence. 

\subsection{Discriminators for Text Generation}
Recent works have applied text classifiers as discriminators for different NLG tasks. \citet{kryscinski2019evaluating} used them to detect factual consistency in the context of abstractive summarization; \citet{zellers2019defending} applied discriminators to detect fake news, in a news generation scenario, reporting high accuracy (over 90\%). Recently, \citet{clark2019electra} proposed to train encoders as discriminators rather than language models, as an alternative to BERT \cite{devlin2019bert}; they obtained better performances while improving in terms of training time.  
Closest to our work, \citet{chen2020discriminator} leverage on discriminators to improve  unconditional text generation following \citet{gabriel2019cooperative} work on summarization.
Abstractive summarization systems tend to be too extractive \cite{kryscinski2018improving}, mainly because of the copy mechanism \cite{vinyals2015pointer}. To improve the abstractiveness of the generated outputs, \citet{gehrmann2018bottom} proposed to train a classifier to detect which words from the input could be copied, and applied it as a filter during inference: to some extent, our work can be seen as the generalisation of this approach.

\subsection{Text Decoding}

\emph{Beam search} is the de-facto algorithm used to decode generated sequences of text, in NLG taks. This decoding strategy allows to select the sequence with the highest probability, offering more flexibility than a greedy approach. Beam search has contributed to performance improvements of state-of-the-art models for many tasks, such as Neural Machine Translation, Summarization, and Question Generation \cite{ott2018analyzing, dong2019unified}. 
However, external rules are usually added to further constrain the generation, like the filtering mechanism for copy described above \cite{gehrmann2018bottom} or the inclusion of a length penalty factor \cite{wu2016google}. \citet{hokamp2017lexically} reported improvements when adding lexical constraints to beam search. Observing that neural models are prone to repetitions, while human-produced summaries contain more than 99\% unique 3-grams,  \citet{paulus2017deep} introduced a rule in the beam forbidding the repetition of 3-grams. 

Whether trained from scratch \cite{paulus2017deep, gehrmann2018bottom} or based on pre-trained language models \cite{dong2019unified}, the current state-of-the-art results in abstractive summarization have been achieved using length penalty and 3-grams repetition avoidance.  

\section{Datasets}
\label{sec:datasets} 

\begin{table}
\centering
\begin{tabular}{lrrr}
    & len\_src & len\_tgt & abstr. (\%) \\
\hline
CNN/DM & 810.69 & 61.04 & 10.23\\
TL;DR  & 248.95 & 30.71 & 36.88              
\end{tabular}
\caption{Statistics of CNN/DM and TL;DR summarization datasets. We report length in tokens for source (len\_src) and summaries (len\_tgt). Abstractiveness (abstr.) is the percentage of tokens in the target summary, which are not present in the source article.}
\label{table:stat_datasets}
\end{table}

While the proposed approach is applicable to any Natural Language Generation (NLG) task, we focus on Abstractive Summarization in this study.
One of most popular datasets for summarization is the CNN/Daily Mail (CNN/DM) dataset \cite{hermann2015teaching, nallapati2016abstractive}. It is composed of news articles paired to multi-sentence summaries. The summaries were written by professional writers and consist of several bullet points corresponding to the important information present in the paired articles.
For fair comparison, we used the exact same dataset version as previous works \cite{see2017get,gehrmann2018bottom,dong2019unified}.\footnote{Publicly available at \url{https://github.com/microsoft/unilm\#abstractive-summarization---cnn--daily-mail}}

Furthermore, to assess the possible benefits of the proposed approach in a domain adaptation setup, we conduct experiments on TL;DR, a large scale summarization dataset built from social media data \cite{volske-etal-2017-tl}.
We choose this dataset for two main reasons: first, its data is relatively out-of-domain if compared to the samples in CNN/DM; second, its characteristics are also quite different: compared to CNN/DM, the TL;DR summaries are twice shorter and three times more abstractive, as detailed in Table \ref{table:stat_datasets}.
The training set is composed of around 3M examples and publicly available,\footnote{\url{https://zenodo.org/record/1168855}} while the test set is kept hidden because of public ongoing leaderboard evaluation. Hence, we randomly sampled 100k examples for training, 5k for validation and 5k for test. For reproducibility purposes, we make the TL;DR split used in this work publicly available.

\section{Discriminative Adversarial Search}
\label{section:das}

The proposed model is composed of a generator $G$ coupled with a sequential discriminator $D$: at inference time, for every new token generated by $G$, the score and the label assigned by $D$ is used to refine the probabilities, within a beam search, to select the top candidate sequences.

While the proposed approach is applicable to any Natural Language Generation (NLG) task, we focus on Abstractive Summarization.

\paragraph{Generator}

Abstractive summarization is usually cast as a sequence to sequence task: 

\begin{equation}
    P_{\gamma }(y | x) = \prod_{t=1}^{|y|} P_{\gamma }(y_t|x, y_{1:t-1}))
    \label{equation:generator}
\end{equation}
where $x$ is the input text, $y$ is the summary composed of $y_1 ... y_{|y|}$ tokens and $\gamma$ represents the parameters of the generator. Under this framework, an abstractive summarizer is thus trained using article ($x$) and summary ($y$) pairs (e.g., via log-likelihood maximization).

\paragraph{Discriminator}
The objective of the discriminator is to label a sequence $y$ as being \textit{human-produced} or \textit{machine-generated}. 
We use the discriminator to obtain a label at each generation step, rather than only for the entire generated sequence. For simplicity, we cast the problem as sequence to sequence, with a slight modification from our generator: at each generation step, the discriminator, instead of predicting the next token among the entire vocabulary $V$, outputs the probability that the input summary was generated by a human.

Learning the neural discriminator  $D_\delta$, 
using parameters $\delta$, corresponds to the following logistic regression problem:  
\begin{equation}
\small 
\frac{1}{|H|} \sum_{(x,y) \in H} \log(D_\delta (x,y)) + \frac{1}{|G|} \sum_{(x,y) \in G} \log(1 - D_\delta (x,y))  
     \label{equation:discr_label}
\end{equation}
where $H$ and $G$ are  sets of  pairs $(x,y)$ of all texts $x\in X$ to be summarized associated to any sub-sequence $y$ (from start to any token index $t$), respectively taken from ground truth summaries and generated ones:
{ 
$$H=\{(x,y_{1:t}) | x \in {X} \wedge y \in H(x) \wedge  t \leq  |y|\}$$ 
$$G=\{(x,y_{1:t}) | x \in {X} \wedge y \in G(x) \wedge  t \leq  |y|\}$$
}
where $x \in {X}$ stands as a text from the training set $X$ and $H(x)$ and $G(x)$ respectively correspond to the associated human-written summary and a set of  generated summaries for text $x$.

We refer to $D_\delta$ as a sequential discriminator, since it learns to discriminate  for any partial sequence (up to the $t$ tokens generated at step $t$) of any summary $y$.
We cut all the summaries to $T=140$ tokens if longer, consistently with previous works~\cite{dong2019unified}.

\subsection{Discriminative Beam Reranking}
\label{subsection:reranking_the_beams}

At inference time, the aim is usually to maximize the probability of the output $y$ according to the generator (Eq.~\ref{equation:generator}). The best candidate sequence 
is the one that maximizes 
$P_\gamma(y|x)$. 
The beam search procedure is a greedy process that iteratively constructs sequences from $y_1$ to $y_n$, while maintaining  a pool of $B$ best hypotheses generated so far at each step  
to allow exploration (when $B>1$). At  each step $t$, the process assigns a score, for every sub-sequence $y_{1:t-1}$ from the pool $B$, to every candidate new token $y_t$ from the vocabulary $V$ : 
\begin{equation}
S_{gen}(\hat{y})=\log P_\gamma(y_{1:t-1}|x) + \log P_\gamma(y_t|x,y_{1:t-1})
\end{equation}
where $\hat{y}$ results from the concatenation of a new token $y_t$ at the end of a sequence $y_{1:t-1}$. The $B$ sequences $\hat{y}$ with best $S_{gen}(\hat{y},x)$ scores are kept to form the pool of hypotheses at next step. 
 Finally, when all sequences from $B$ are ended sequences (with the end token $\$$ as the last token $\hat{y}_{-1}$), the one with best $S_{gen}$ score is returned.  The beam size $B$ corresponds to a hyper-parameter which enables to control exploration and complexity of the process. It ranges between 1 and 5 in the literature.

\begin{algorithm}
\caption{DAS: a Beam Search algorithm with the proposed discriminator re-ranking mechanism highligted.}
\label{alg:beam_reranking}
\begin{algorithmic}[1]
\REQUIRE $B$, $T$, $K_{rerank}$, $\alpha$ 
\STATE $C\leftarrow\{$Start-Of-Sentence$\}$
\FOR{$t=1,...,T$}
\STATE $C \leftarrow \{\hat{y} | (\hat{y}_{1:t-1} \in C \wedge \hat{y}_{t} \in V)$\\ $\ \ \   \qquad \qquad \vee (\hat{y} \in C \wedge \hat{y}_{-1}=\$)\}$

\ 

\vspace{-0.3cm}
\#        Pre-filter  $K_{rerank}$ sequences with top $S_{gen}$
\STATE $C \leftarrow \argmax\limits_{\tilde{C} \subseteq C, |\tilde{C}|=K_{rerank}} \sum_{\hat{y} \in \tilde{C}} S_{gen}(\hat{y})$

\tikzmk{A}

\#        Filter  $B$ sequences with top 
$S_{DAS}$
\STATE $C \leftarrow \argmax\limits_{\tilde{C} \subseteq C, |\tilde{C}|=B} \sum_{\hat{y} \in \tilde{C}} S_{DAS}(\hat{y})$

\tikzmk{B}\boxit{pink}


\ 

\IF{only ended sequences in $C$} 
\STATE \textbf{return} $C$
\ENDIF
\ENDFOR
\end{algorithmic}
\end{algorithm}






In our method, we propose a new score $S_{DAS}$ to refine the score $S_{gen}$ during the beam search w.r.t. the log probability of the discriminator, such that:
\begin{equation}
    S_{DAS}(\hat{y}) = S_{gen}(\hat{y}) + \alpha \times 
    S_{dis}(\hat{y}) \label{equation:S_DAS}
\end{equation}
where $S_{dis}(\hat{y})=log(D_\delta( x, \hat{y})$ 
is the discriminator 
log-probability that the sequence $\hat{y}$ is human-written; 
$\alpha \geq 0$ is used as a weighting factor. 
While theoretically such scores could be computed for the entire vocabulary, in practice  applying the discriminator to all of the $|V| \times B$ candidate sequences $\hat{y}$ 
at every step $t$ 
would be too time-consuming. For complexity purposes, we thus 
limit the re-ranking to the pool
of the $K_{rerank}$ sequences with best $S_{gen}(\hat{y},x)$ score, 
as detailed in Algorithm~\ref{alg:beam_reranking}.

\subsection{Retraining the Discriminator}

\begin{figure}
\centering\includegraphics[width = \columnwidth]{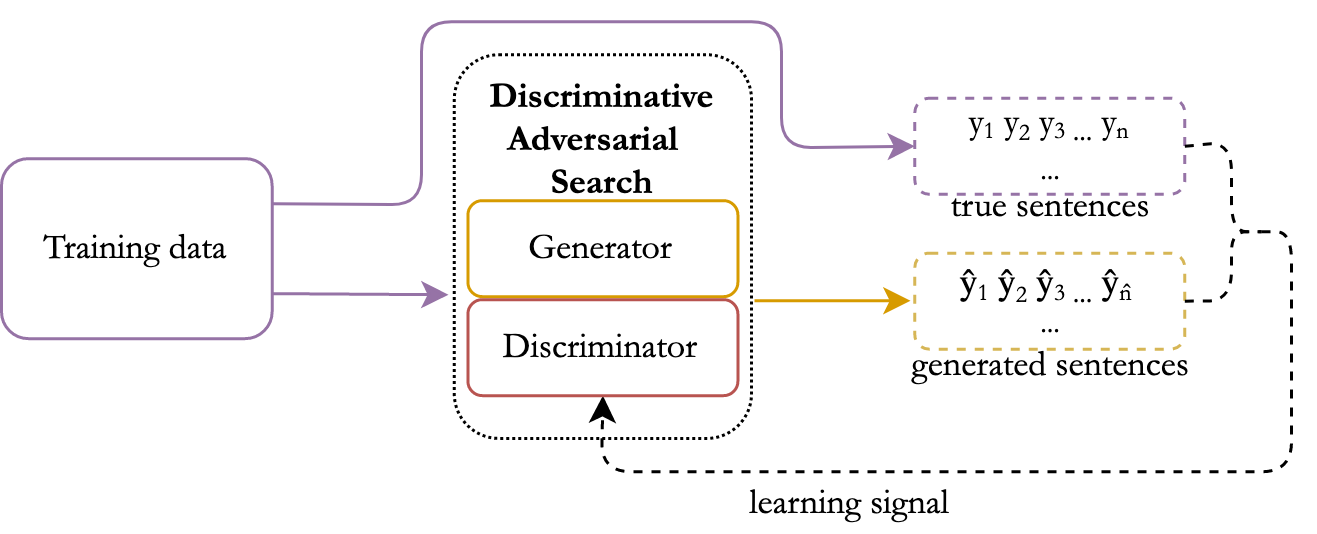}
\caption{DAS self-training procedure: the generated examples are improved by the discriminator, and then fed back to the discriminator in a self-training loop. 
}
\label{fig:retraining_discriminator}
\end{figure}

Under the proposed paradigm, as mentioned in Section~\ref{section:introduction}, the discriminator can be fine-tuned using the outputs  generated
from the re-ranking process, to match the new generation distribution.
Inspired by the GAN paradigm, we iteratively retrain the discriminator given the new predictions until convergence. We detail the full procedure in Figure \ref{fig:retraining_discriminator}, where  
 the discriminator is retrained iteratively following equation \ref{equation:discr_label} at each step. The set of  generated summaries $G$ used in equation \ref{equation:discr_label} to train the discriminator at step $t+1$  corresponds to outputs 
 from our DAS process using the discriminator at step $t$. This allows to consider at each step a discriminator that attempts to correct output distributions from the previous step, in order to incrementally converge to realistic distributions (w.r.t. human summaries), without requiring any retraining of the generator. 

\section{Experimental Protocol}

\paragraph{Generator} 
We build upon the Unified Language Model for natural language understanding and generation (UniLM) proposed by \citet{dong2019unified}; it is the current state-of-the-art model for summarization.\footnote{Code and models available at \url{https://github.com/microsoft/unilm\#abstractive-summarization---cnn--daily-mail}} 
This model can be described as a Transformer \cite{vaswani2017attention} whose weights are first initialised from BERT. However, BERT is an encoder trained with bi-directional self attention: it can be used in Natural Language Understanding (NLU) tasks but not directly for generation (NLG). \citet{dong2019unified} proposed to unify it for NLU and NLG: resuming its training, this time with an unidirectional loss; after this step, the model can be directly fine-tuned on any NLG task.

For our ablation experiments, to save time and computation, we do not use UniLM (345 million parameters). Instead, we follow the approach proposed by the authors \cite{dong2019unified}, with the difference that \emph{1)} we start from BERT-base (110 million parameters) and \emph{2)} we do not extend the pre-training. We actually observed little degradation than when starting from UniLM. We refer to this smaller model as \emph{BERT-gen}. 
For our final results we instead use the exact same UniLM checkpoint made publicly available by \citet{dong2019unified} for Abstractive Summarization. 

\paragraph{Discriminator} As detailed in Section~\ref{section:das}, the discriminator model is also based on a sequence to sequence architecture. Thus, we can use again \emph{BERT-gen}, initializing it the same way as the generator. 
The training data from CNN/DM is used to train the model; for each sample, the discriminator has access to two training examples: the human reference
and a generated summary.

Hence, the full training data available for the discriminator amounts to $\sim600k$ total examples. However, as detailed in the following Section, the discriminator does not need a lot of data to achieve a high accuracy. Therefore, we only used 150k training examples, split into 90\% for training, 5\% for validation and 5\% for test. Unless otherwise specified, this data is only used to train/evaluate the discriminator.

\paragraph{Implementation details}
All models are implemented in PyText \cite{aly2018pytext}. 
For all our experiments we used a single RTX 2080 Ti GPU.

To train the discriminator, we used the Adam optimiser with the recommended parameters for BERT: learning rate of  $3e^{-5}$, batch size of 4 and accumulated batch size of 32. We trained it for 5 epochs; each epoch took 100 minutes on 150k samples. 

During discriminator retraining, the generator is needed and thus additional memory is required: all else equal, we decreased the batch size to 2. The self-training process takes one epoch to converge, in about 500 minutes: 200 minutes for training the discriminator and 300 minutes to generate the summaries with the search procedure described in Algorithm~\ref{alg:beam_reranking}.

\paragraph{Metrics}

The evaluation of NLG models remains an open research question. Most of the previous works report n-grams based metrics such as ROUGE \cite{lin-2004-rouge} or BLEU \cite{papineni2002bleu}. 
ROUGE-n is a recall oriented metric counting the percentage of n-grams in the gold summaries that are present in the evaluated summary. Conversely, BLEU is precision oriented. 

However, as discussed in Section~\ref{section:introduction}, these metrics do not correlate sufficiently w.r.t human judgments. For summarization, \citet{louis2013automatically} showed how this issue gets even more relevant when few gold references are given. Unfortunately, the annotation of large scale dataset is not realistic: in practice, all the large scale summarization datasets rely on web-scraping to gather text-summary pairs. 

For these reasons, \citet{see2017get} suggested to systematically compare summarization systems with other metrics such as novelty and the number of repetitions.
Following the authors' recommendation, we report the following measures for all our experiments: \emph{i)} Novelty (\emph{nov-n}), as the percentage of novel n-grams w.r.t. the source text, indicating the abstractiveness of a system; \emph{ii)} Repetition (\emph{rep-n}), as the percentage of n-grams that occur more than once in the summary; and \emph{iii)} Length (\emph{len}), as the length in tokens of the summary.

It is important to note that the objective is not to maximize those metrics, but to minimize the difference w.r.t. human-quality summaries. Hence, we report this difference such that for any measure $m$ above, $\Delta m = m_{human} - m_{model}$.

\section{Preliminary study}

\begin{figure}\centering\includegraphics[width =  \columnwidth]{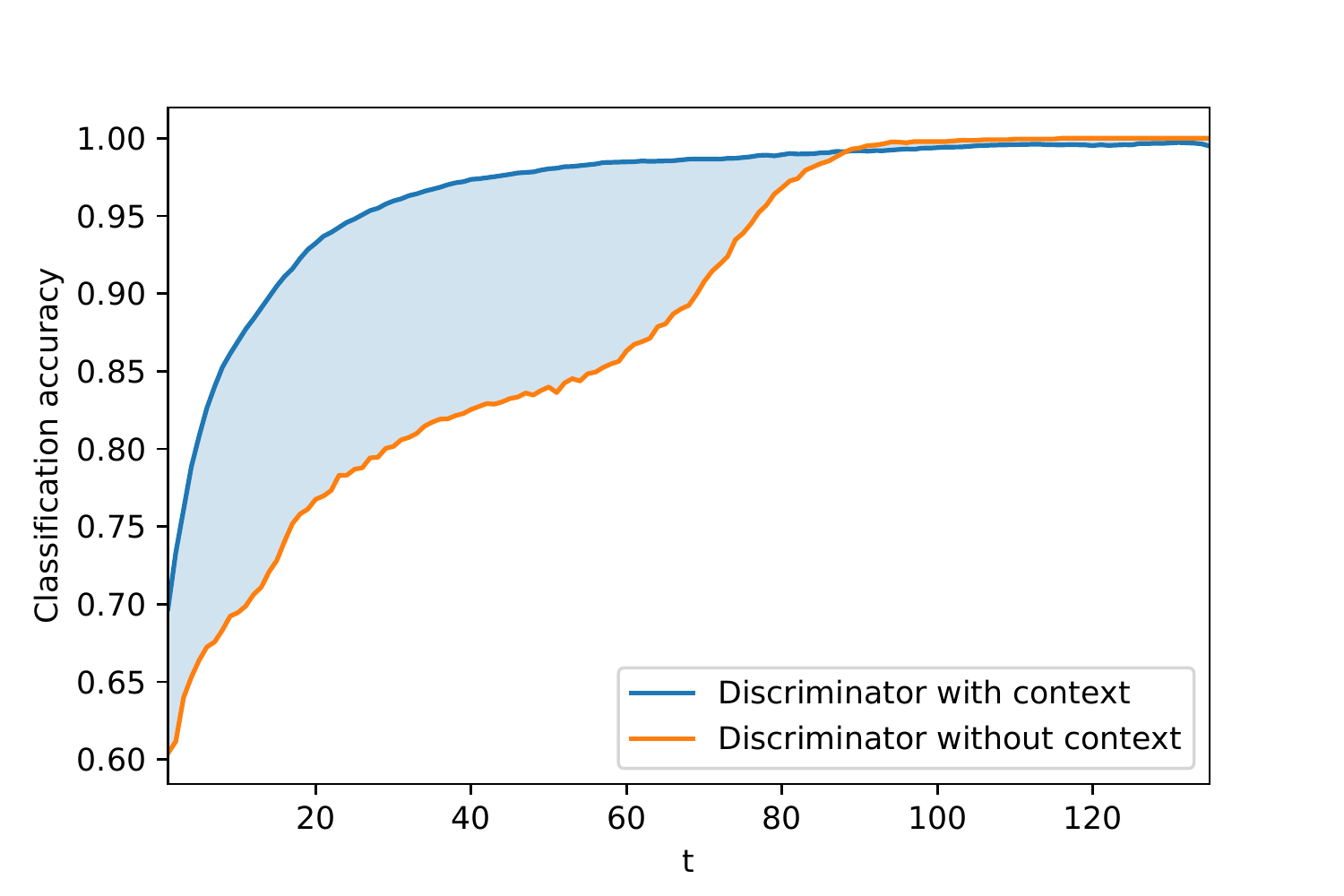}
\caption{Accuracy of two discriminators: one is given access to the source context $x$ while the other is not.  The x-axis corresponds to the length of the discriminated sub-sequences.}
\label{fig:accuracy_discriminator_per_length}
\end{figure}

High discriminator accuracy is of utmost importance for DAS to improve the decoding search. In Fig.~\ref{fig:accuracy_discriminator_per_length} we plot the discriminator accuracy against the generation step $t$, with $t$ corresponding to the prediction for the partial sequence of the summary, $y_1, ..., y_t$ (see Eq.~\ref{equation:discr_label}). As an ablation, we report the accuracy for a discriminator which is not given access to the source article $x$. 

As one would expect, the scores improve with higher $t$, from 65\% for $t=1$ to 98\% for $t=140$: the longer the sequence $y_1, ..., y_t$ of the evaluated summary, the easier it is to discriminate it. This observed high accuracy indicates the potential benefit of using the discriminator signal to improve the generated summaries.

When trained without access to the source article $x$ (orange plot), 
the discriminator has access to little contextual and semantic information and its accuracy is lower than a discriminator who has access to $x$.
In Fig.~\ref{fig:accuracy_discriminator_per_length}, the shaded area between the two curves represents the discrimination performance improvement attributed to using the source article $x$. It increases for $1 \leq t \leq 40$ and starts shrinking afterwards. After $t=60$, corresponding to the average length of the human summaries (see Table~\ref{table:stat_datasets}), the performance of the discriminator without context quickly increases, indicating that the generated sequences contain relatively easy-to-spot mistakes. This might be due to the increased difficulty for the generator to produce longer and correct sequences, as errors may accumulate over time.

\paragraph{Impact of $K_{rerank}$ and $\alpha$}
\begin{table}[]
\centering
\begin{tabular}{ll rrr}
& $K_{rerank}$  & DAS-single & DAS-retrain \\
\toprule
 & 1 (BERT-gen) & 27.70$\pm$0.3  & 27.70$\pm$0.3 \\
\smash{\rotatebox[origin=c]{90}{BLEU-1}}  & 5 & 27.51$\pm$0.3 & 29.70$\pm$0.3 \\
 & 10 & 29.18$\pm$0.3 & 29.81$\pm$0.2 \\
&   &  & \\

 & 1 (BERT-gen) & 11.71$\pm$0.1 & 11.71$\pm$0.1 \\
\smash{\rotatebox[origin=c]{90}{$\Delta$ nov-1}}  & 5 & 11.22$\pm$0.1 & 10.05$\pm$0.2 \\
& 10 & 10.83$\pm$0.3 & 9.82$\pm$0.1  \\
&   &  & \\

 & 1 (BERT-gen) & -9.84$\pm$0.1 & -9.84$\pm$0.1 \\
\smash{\rotatebox[origin=c]{90}{$\Delta$ len}}  & 5 & -7.24$\pm$0.1 & -3.05$\pm$0.1 \\
& 10 & -3.14$\pm$0.1 & -1.42$\pm$0.1   \\
&   &  & \\

 & 1 (BERT-gen) & -21.49$\pm$1.2 & -21.49$\pm$1.2 \\
 \smash{\rotatebox[origin=c]{90}{$\Delta$ rep-3}} & 5 & -17.54$\pm$0.5 & -11.26$\pm$0.4 \\
 & 10 & -13.77$\pm$0.8 & -10.45$\pm$0.4    

\end{tabular}
\caption{Scores obtained with varying $K_{rerank}$}
\label{table:ablation_K_rerank}
\end{table}

To assess the behavior of DAS, we conducted experiments with \emph{BERT-gen} for both the generator and the discriminator using different values for $\alpha$ and $K_{rerank}$. All models are trained using the same training data from CNN/DM, and the figures reported in Tables~\ref{table:ablation_K_rerank} and~\ref{tab:alpha} are the evaluation results averaged across 3 runs on three different subsets (of size 1k) randomly sampled from the validation split.
We compare  \emph{i)} \emph{BERT-gen}, \emph{i.e.} the model without a discriminator, \emph{ii)} DAS-single, where the discriminator is not self-retrained, and \emph{iii)} DAS-retrain, where the discriminator is iteratively retrained. As previously mentioned, for the repetition, novelty and length measures, we report the difference w.r.t. human summaries: the closer to 0 the better -- 0 indicates no difference w.r.t. human. 

The parameter $K_{rerank}$ corresponds to the number of explored possibilities by the discriminator (see Sec.~\ref{subsection:reranking_the_beams}). With $K_{rerank}=1$, no reranking is performed, and the model is equivalent to \emph{BERT-gen}. 

In  Table~\ref{table:ablation_K_rerank}, for which we set $\alpha=0.5$, 
we observe that both increasing $K_{rerank}$ and retraining the discriminator help to better fit the human distribution (\emph{i.e.} lower $\Delta$): compared to \emph{BERT-gen}, DAS models generate more novel words, are shorter and less repetitive, show improvements over the base architecture, and also obtain performance gains in terms of BLEU.

Further, we report in Table~\ref{tab:alpha} results for DAS models with a fixed  $K_{rerank}=10$, while varying $\alpha$. 
$\alpha$ controls the impact of the discriminator predictions when selecting the next token to generate (see Eq.~\ref{equation:S_DAS}). 

With $\alpha=0$, the discriminator is deactivated and only the generator probabilities $S_{gen}$ are used (corresponding to Eq.~\ref{equation:generator}): the model is effectively equivalent to \emph{BERT-gen}. Consistently with the results obtained for varying $K_{rerank}$, we observe: DAS-retrain $>$ DAS-single $>$ \emph{BERT-gen} for $\alpha \neq 5$. However, when $\alpha=5$, 
BLEU scores decrease.
This could indicate that a limit was reached: the higher the $\alpha$, the more the discriminator influences the selection of the next word. 

With $\alpha=5$, the generated sequences are too far from the generator top-p probabilities, selected tokens at step $t$ do not lead to useful sequences in the best $K_{rerank}$ candidates at the following steps. The generation process 
struggles to represent sequences too far from what was seen during training.

\begin{table}[]
\centering
\begin{tabular}{ll rr}
& $\alpha$ & DAS-single & DAS-retrain \\
\toprule
& 0 (BERT-gen) & 27.70$\pm$0.3 & 27.70$\pm$0.3  \\
\smash{\rotatebox[origin=c]{90}{BLEU-1}}  & 0.5 & 27.51$\pm$0.3 & 29.70$\pm$0.3 \\
& 1 & 28.38$\pm$0.3 & 29.25$\pm$0.2 \\
& 5 & 24.26$\pm$0.4 & 27.47$\pm$0.4 \\
&   &  & \\

 & 0 (BERT-gen) & 11.71$\pm$0.1 & 11.71$\pm$0.1  \\
\smash{\rotatebox[origin=c]{90}{$\Delta$ nov-1}}  & 0.5 & 11.22$\pm$0.1 & 10.05$\pm$0.2 \\
& 1 & 10.70$\pm$0.2 & 9.33$\pm$0.1  \\
& 5 & 7.98$\pm$0.2 & 6.57$\pm$0.2 \\

&   &  & \\

 & 0 (BERT-gen) & -9.84$\pm$0.1 & -9.84$\pm$0.1 \\
\smash{\rotatebox[origin=c]{90}{$\Delta$ len}}  & 0.5 & -7.24$\pm$0.1 & -3.05$\pm$0.1 \\
& 1 & -4.11$\pm$0.1 & -3.10$\pm$0.1   \\
& 5 & -7.11$\pm$0.1 & -3.85$\pm$0.1  \\

&   &  & \\

& 0 (BERT-gen)  & -21.49$\pm$1.2 & -21.49$\pm$1.2  \\
\smash{\rotatebox[origin=c]{90}{$\Delta$ rep-3}}  & 0.5 & -17.54$\pm$0.5 & -11.26$\pm$0.4 \\
& 1 & -12.85$\pm$0.4 & -8.93$\pm$0.4\\
& 5 & -2.19$\pm$0.3 & -5.49$\pm$0.3

\end{tabular}

\caption{Scores obtained with varying $\alpha$}
\label{tab:alpha}
\end{table}










\section{Results and discussion}
\begin{table*}[!h]
\centering

\begin{tabular}{lrrrrrrrr}
                 & $\Delta$ len  & $\Delta$ nov-1 & $\Delta$ nov-3 & $\Delta$ rep-1 & $\Delta$ rep-3 & R1 & RL & B1 \\
\hline
\citeauthor{see2017get}  & - & -    &-    & -  & -  & 36.38  & 34.24  & -\\ 
\citeauthor{gehrmann2018bottom}  & - & -   & -  & -  & -  & 41.22  & 38.34  & -\\ 
\citeauthor{kryscinski2018improving}  & - & 10.10    & 32.84  & -  & -  & 40.19  & 37.52  & -\\ 
UniLM            & -40.37 & 8.35    & 7.98    & -27.99   & \textbf{0.12}   & 43.08   & 40.34   & 34.24  \\
UniLM (no rules) & -45.57 & 8.58    & 7.98    & -31.41   & -6.88    & 42.98   & 40.54   & 34.46  \\
DAS-single & -29.75 & \textbf{6.05}    & 2.80    & -28.21   & -4.60    & 42.90   & 40.05   & 35.69  \\
DAS-retrain              & \textbf{-16.81} & 6.69    & \textbf{2.59}    & \textbf{-25.21}   & -2.40    & \textbf{44.05}   & \textbf{40.58}   & \textbf{35.94} 
\end{tabular}
\caption{Results on CNN/DM test set for the previous works as well as our proposed models.}
\label{table:cnn_main_result}
\end{table*}

\begin{table*}[!h]
\centering
\begin{tabular}{lrrrrrrrr}
                 & $\Delta$ len  & $\Delta$ nov-1 & $\Delta$ nov-3 & $\Delta$ rep-1 & $\Delta$ rep-3 & R1 & RL & B1 \\
\hline
UniLM            & -12.11 & 27.16   & 5.49    & -6.87    & \textbf{0.19}   & 18.66   & \textbf{15.49}   & 16.91  \\
UniLM (no rules)  & -13.11 & 30.16   & 5.69    & -7.87    & -3.77    & 18.76   & 14.49   & 17.14  \\
DAS-single & -10.76 & 19.68   & 4.58    & -10.81   & -5.05    & 18.19   & 13.30   & 15.41  \\
DAS-retrain              & \textbf{-2.72}  & \textbf{19.05}   & \textbf{1.01}    & \textbf{-3.42}    & -1.33    & \textbf{19.76}   & 14.92   & \textbf{17.59}
\end{tabular}
\caption{Results on TL;DR test set for our proposed model in transfer learning scenarios.}
\label{table:tldr_main_result}
\end{table*}


In our preliminary study, the best performing DAS configuration was found with $K_{rerank}=10, \alpha=1$. 
We apply this configuration in our main experiments, for fair comparison using the state-of-the-art UniLM model checkpoint.\footnote{As publicly released by the authors.} Results on the CNN/DM test set are reported in Table~\ref{table:cnn_main_result}. 
Confirming our preliminary study, DAS favorably compares to previous works, for all the metrics.
Compared to UniLM, we can observe that both DAS-single and DAS-retrain are closer to the target data distribution: they allow to significantly reduce the gap with human-produced summaries over all metrics.
The length of the summaries are 16.81 tokens in average longer than the human, as opposed to 40.37 tokens of difference for UniLM and 45.57 without the length penalty. DAS-retrain is also more abstractive, averaging only 2.59 novel 3-grams less than the human summaries, as opposed to 7.98 for UniLM. Notably, the proposed approach also outperforms \citet{kryscinski2018improving} in terms of novelty, while their model was trained with novelty as a reward in a reinforcement learning setup.
UniLM applies a 3-grams repetition avoidance rule, which is why this model generates even less 3-grams repetitions than human summaries. Without this post-hoc rule, DAS-retrain generation is less repetitive compared to UniLM.
Incidentally, our approach also outperforms the previous works and achieves, to the best of our knowledge, a new state-of-the-art for ROUGE. 
\begin{figure}
\centering
\includegraphics[width=\columnwidth]{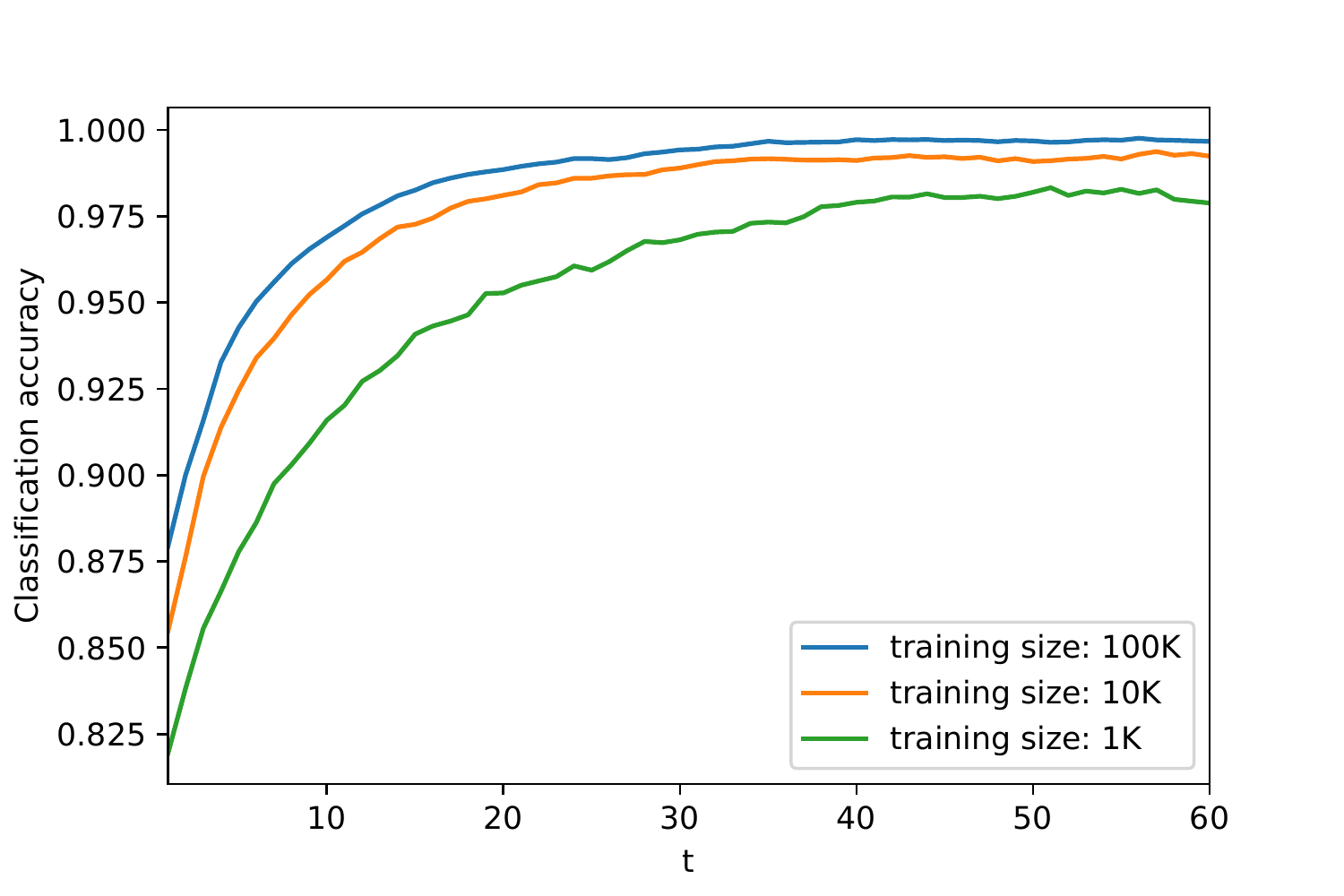}
\caption{Learning curve for discriminators trained on TL;DR on 1k, 10k and 100k examples. The x-axis corresponds to the length of the discriminated sub-sequences.} 
\label{fig:learningcurve_disc_tldr}
\end{figure}
\begin{figure*}
\centering
\includegraphics[width = \columnwidth]{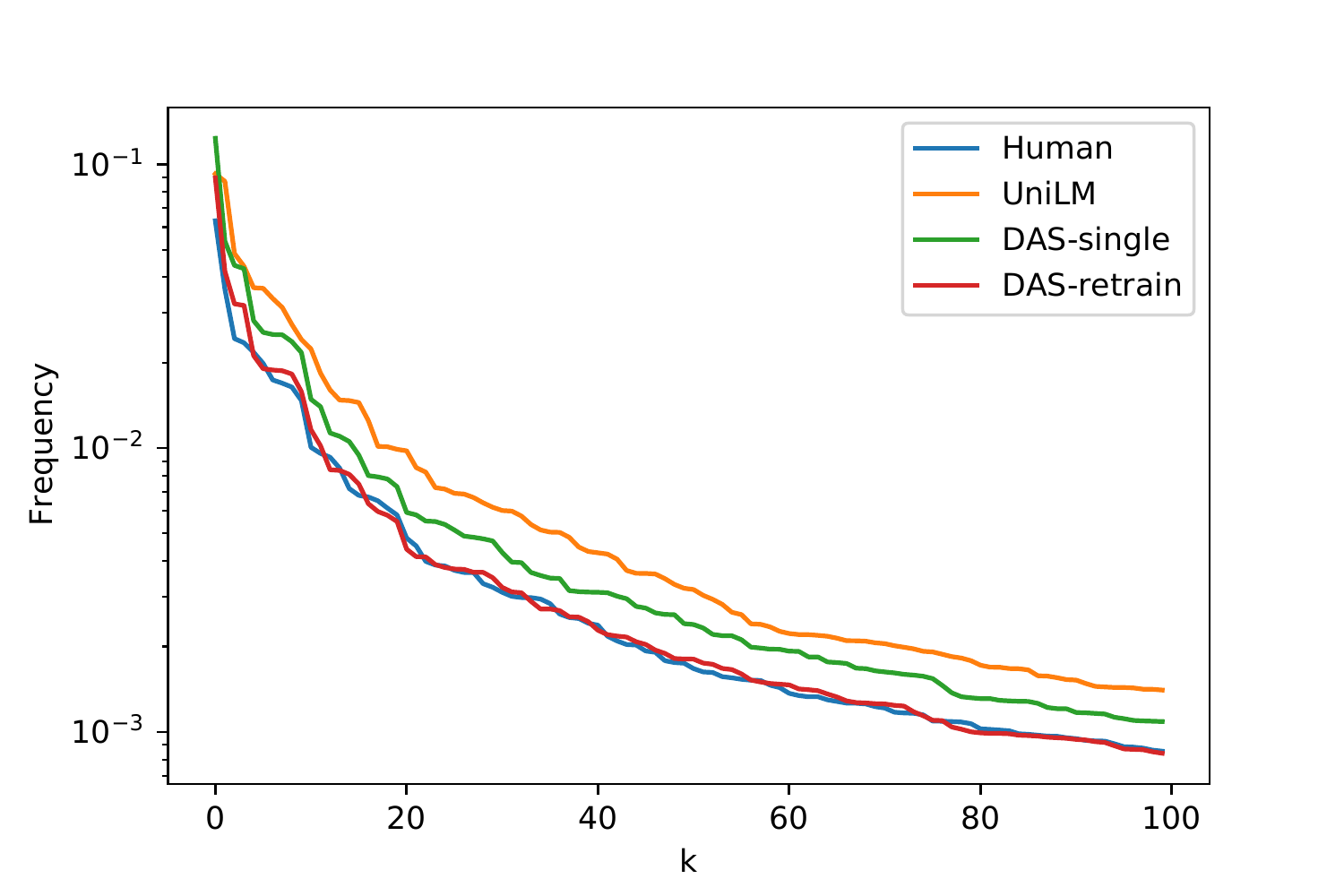}
\includegraphics[width = \columnwidth]{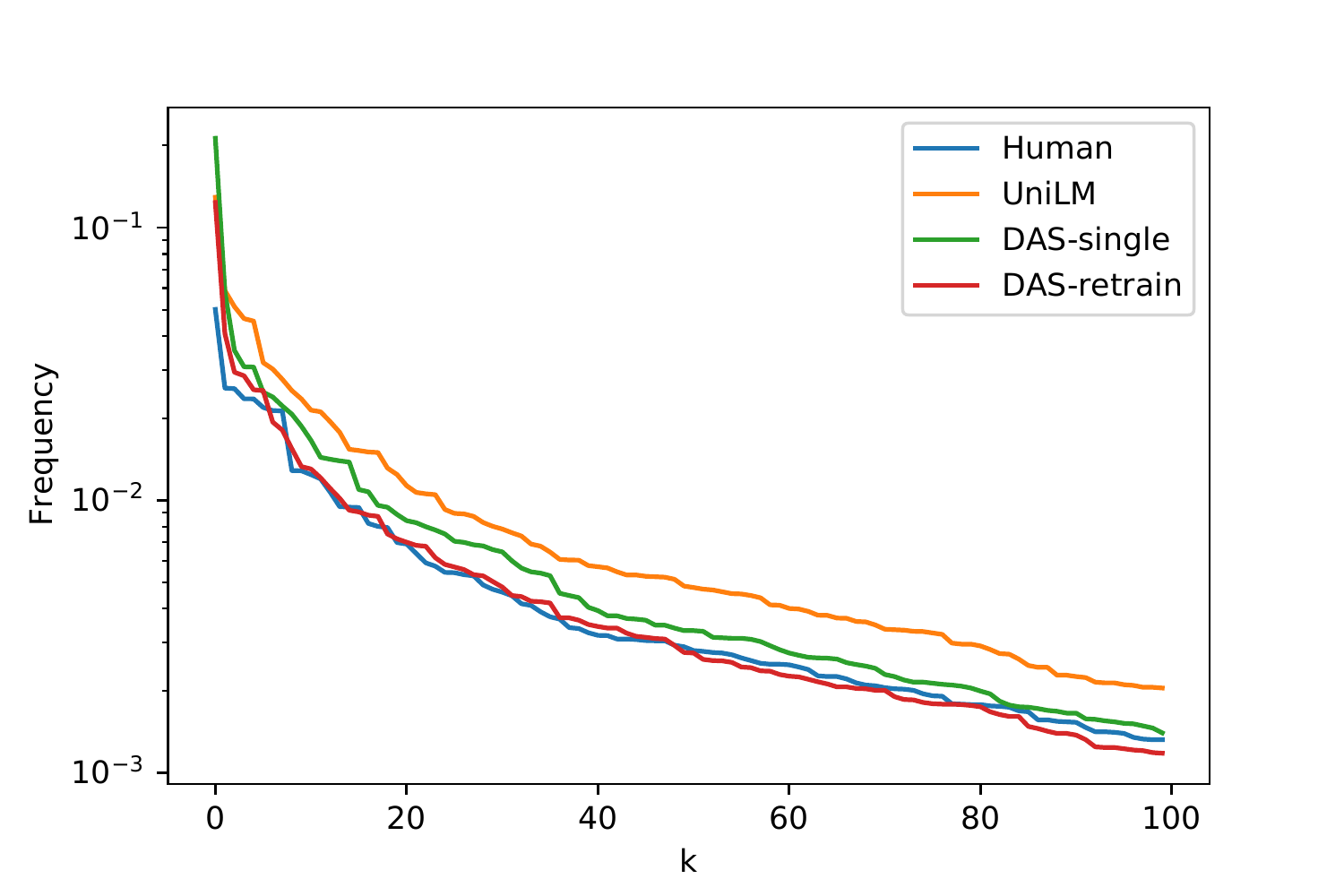}
\caption{Vocabulary frequency for the $k=100$ most frequent words generated by the models, for CNN/DM (left) and TL;DR (right).}
\label{fig:zip_cnn}
\end{figure*}

\begin{figure}
\centering
\includegraphics[width = \columnwidth]{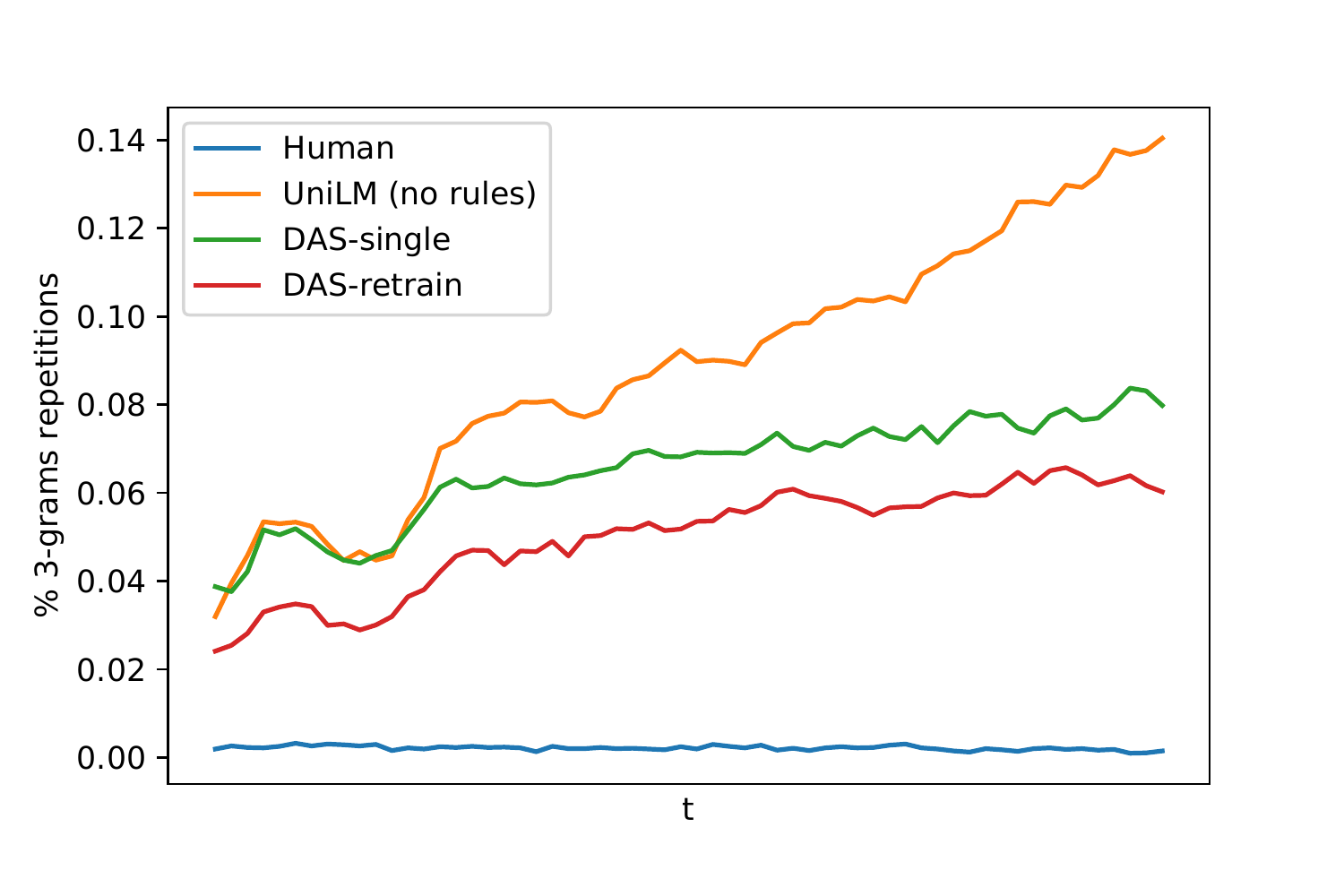}
\caption{Distribution of 3-grams repetitions over their position $t$ in the sequence (CNN/DM data).}
\label{fig:dist_3grams}
\end{figure}

\paragraph{Domain Adaptation}

Further, in Fig.~\ref{table:tldr_main_result}, we explore a domain adaptation scenario, applying DAS-retrain on a second dataset, TL;DR. This dataset is built off social media data, as opposed to news articles as in CNN/DM, and differs from the latter in several respects, as described in Section~\ref{sec:datasets}.
In this scenario, we keep the previously used generator (\emph{i.e.} the UniLM checkpoint trained on CNN/DM), and only train the discriminator on a subset of TL;DR training samples. 
This setup has practical applications in scenarios where limited data is available: indeed, learning to generate is harder than to discriminate and requires a large amount of examples \cite{gehrmann2018bottom}. A discriminator can be trained with relatively few samples: in Fig.~\ref{fig:learningcurve_disc_tldr} we show the learning curves for discriminators trained from scratch on TL;DR training subsets of varying size. The samples are balanced: a training set size of 10k means that 5k gold summaries are used, along with 5k generated ones.
We observe that only 1k examples allow the discriminator to obtain an accuracy of 82.5\% at step $t=1$. This score, higher in comparison to the one obtained on CNN/DM (see Fig.~\ref{fig:accuracy_discriminator_per_length}) is due to the relatively lower quality of the \emph{out-of-domain} generator outputs, which makes the job easier for the discriminator.

The results on TL;DR\footnote{Models participating to the public TL;DR leaderboard (\url{https://tldr.webis.de/}) are omitted here, since they are trained on TL;DR data, and evaluated on a hidden test set.
Nonetheless, assuming that the distribution of our sampled test set is similar to that of the official test set, we observe that our approach obtains comparable performance to the state-of-the-art, under a domain-adaptation setup and using only 1k training examples -- exclusively for the discriminator -- over an available training set of 3M examples.} (Table~\ref{table:tldr_main_result}) show larger improvements of DAS-retrain over UniLM, than on CNN/DM.
Due to the high accuracy of the discriminator, the summaries generated are only 2.72 tokens shorter than the human ones as opposed to 12.11. They also contain more novelty and less repetitions.
In terms of ROUGE and BLEU, DAS-retrain also compares favorably with the exception of ROUGE-L.
This might be due to the shorter length of DAS-retrain summaries as compared to UniLM: ROUGE is a recall-oriented metric and ROUGE-L is computed for the longest common sub-sequence w.r.t. the ground truth.

\paragraph{Discussion}

In Fig.~\ref{fig:zip_cnn} we report the frequency distributions for the different models and the human summaries. 
We observe that DAS-retrain comes closer to the human distribution, followed by DAS-single and significantly outperforming UniLM. This shows the benefit of DAS at inference time, to produce relatively more human-like summaries.
Further, the distribution of 3-grams repetition across their relative position in the sequence -- Fig.~\ref{fig:dist_3grams}  -- shows how the gap between UniLM and Human increases more than that between DAS-retrain and human, indicating that our approach contributes to reduce the exposure bias effect. 
Rather than exclusively targeting exposure bias (as in Scheduled Sampling or Professor Forcing), or relying on automatic metrics as in reinforcement learning approaches, we optimize towards a discriminator instead of discrete metrics: besides reducing the exposure bias issue, this allows improvements over the other aspects captured by a discriminator.

\section{Conclusion}
We introduced a novel sequence decoding approach, which directly optimizes on the data distribution rather than on external metrics. 

Applied to Abstractive Summarization, the distribution of the generated sequences are found to be closer to that of human-written summaries over several measures, while also obtaining improvements over the state-of-the-art. 

We reported extensive ablation analyses, and showed the benefits of our approach in a domain-adaptation setup. Importantly, all these improvements are obtained without any costly generator retraining. 
In future work, we plan to apply DAS to other tasks such as machine translation and dialogue systems.

\bibliography{example_paper}
\bibliographystyle{icml2020}

\end{document}